\documentclass[10pt,twocolumn,letterpaper]{article}

\usepackage{cvpr}
\usepackage{times}
\usepackage{epsfig}
\usepackage{graphicx}
\usepackage{amsmath}
\usepackage{amssymb}
\usepackage{subfig}
\usepackage{algorithmic}
\usepackage{tablefootnote}
\usepackage[ruled,vlined]{algorithm2e}

\SetKwInput{KwStage}{Stage}
\SetKwInput{KwInput}{Input}
\SetKwInput{KwOutput}{Output}

\usepackage[pagebackref=true,breaklinks=true,letterpaper=true,colorlinks,bookmarks=false]{hyperref}

\cvprfinalcopy 

\def\cvprPaperID{****} 

\ifcvprfinal\fi
\begin{document}

\title{An Efficient Approach for Anomaly Detection in Traffic Videos}

\author{Keval Doshi\\
University of South Florida\\
4202 E Fowler Ave, Tampa, FL 33620\\
{\tt\small kevaldoshi@mail.usf.edu}
\and
Yasin Yilmaz\\
University of South Florida\\
4202 E Fowler Ave, Tampa, FL 33620\\
{\tt\small yasiny@usf.edu}
}

\maketitle

\begin{abstract}
Due to its relevance in intelligent transportation systems, anomaly detection in traffic videos has recently received much interest. It remains a difficult problem due to a variety of factors influencing the video quality of a real-time traffic feed, such as temperature, perspective, lighting conditions, and so on. Even though state-of-the-art methods perform well on the available benchmark datasets, they need a large amount of external training data as well as substantial computational resources. In this paper, we propose an efficient approach for a video anomaly detection system which is capable of running at the edge devices, e.g., on a roadside camera. The proposed approach comprises a pre-processing module that detects changes in the scene and removes the corrupted frames, a two-stage background modelling module and a two-stage object detector. Finally, a backtracking anomaly detection algorithm computes a similarity statistic and decides on the onset time of the anomaly. We also propose a sequential change detection algorithm that can quickly adapt to a new scene and detect changes in the similarity statistic. Experimental results on the Track 4 test set of the 2021 AI City Challenge show the efficacy of the proposed framework as we achieve an F1-score of 0.9157 along with 8.4027 root mean square error (RMSE) and are ranked fourth in the competition.    
\end{abstract}

\section{Introduction}

\begin{figure}
\centering
\subfloat{\includegraphics[width=3.9cm]{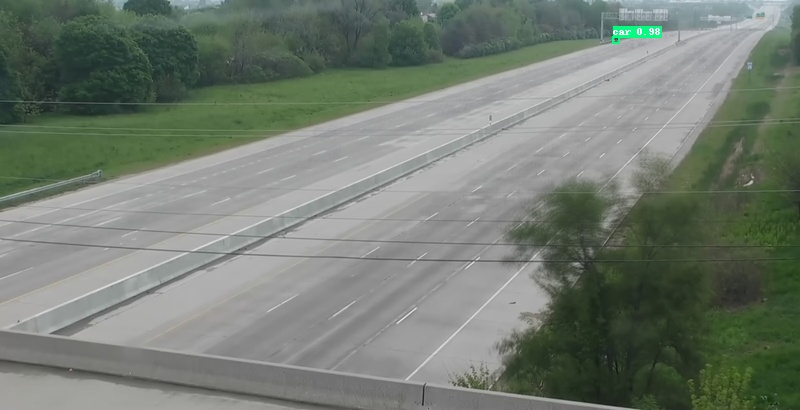}}
\qquad
\hspace{-4mm}
\subfloat{\includegraphics[width=3.9cm]{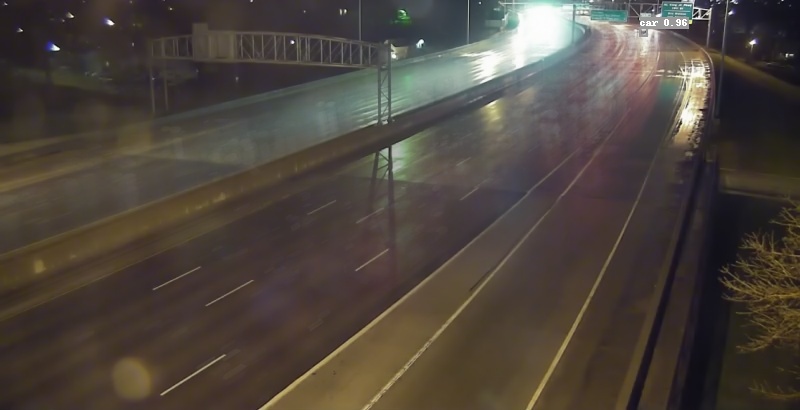}}
\qquad
\subfloat{\includegraphics[width=3.9cm]{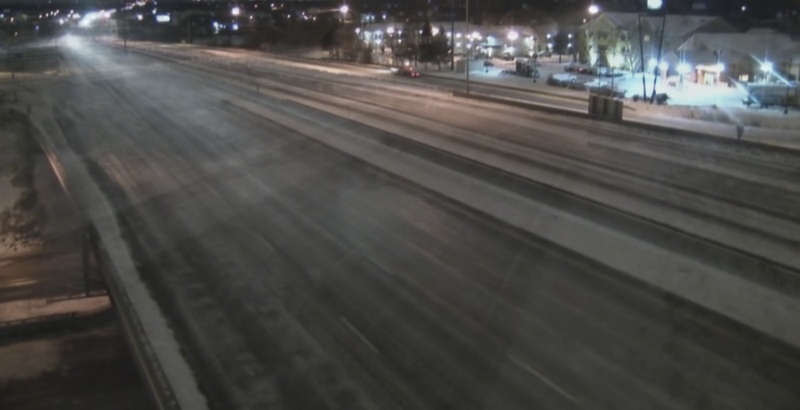}}
\qquad
\hspace{-4mm}
\subfloat{\includegraphics[width=3.9cm]{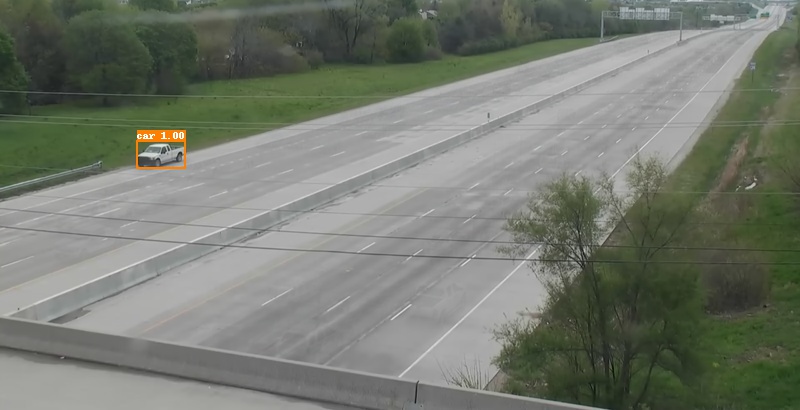}}
\caption{Challenging scenarios from the test videos. Several anomalies in the test videos are not obvious and difficult to detect by humans as well.}

\label{f:challenge}
\vspace{-2mm}
\end{figure}

The identification of abnormal events such as traffic collisions, violations, and crimes is one of the most crucial, demanding, and time-critical tasks in automated traffic video monitoring. As a result, video anomaly detection has become a subject of increasing interest, thanks to its applications in intelligent transportation systems. Anomaly detection is a broad, important, and difficult research subject that deals with identifying data instances that deviate from nominal trends, as shown in Fig. \ref{f:challenge}.

Given the critical role that video anomaly detection will play in ensuring security, stability, and in some cases, the avoidance of possible disasters, one of the most valuable features of a video anomaly detection system is the ability to make real-time decisions. Traffic collisions, robberies, and fires in remote locations necessitate urgent countermeasures, which can be aided by real-time anomaly identification. Despite their relevance, online and real-time detection approaches have received only a small amount of research. Regarding the importance of timely detection in video, as \cite{mao2019delay} argues, the methods should also be evaluated in terms of the average delay, in addition to the commonly used metrics such as true positive rate, false positive rate, and AUC.

Traditionally, the video anomaly detection problem has been formulated as detecting patterns which are previously unseen in the training data. Most of the recent video anomaly detection methods are based on end-to-end trained complex deep neural networks \cite{sultani2018real,liu2018future}, which require a significant amount of data to train. However, a majority of such approaches can only work on datasets from homogeneous scenes, which is not the case in traffic datasets \cite{li2020multi}. Moreover, such models need to be specifically trained on videos from each scene it is deployed at, which is not ideal for edge applications.

To tackle the above challenges, we propose a vehicle detection and background modelling based approach for traffic video anomaly detection. We hypothesize that whenever a traffic accident occurs or a vehicle stops abruptly, the vehicle would be stationary in the video for a significant period of time. To this end, we propose using a Gaussian Mixture Model (GMM) to first eliminate the moving vehicles. Then, we use a pretrained semantic segmentation model to eliminate vehicles in parking lots and some false detections. Since the segmentation model needs to be run only once, we do not include its computational overhead. In an ideal scenario, all vehicles from the videos except for stalled vehicles on/near roads would be removed. We then use a vehicle detection model to extract such stalled vehicles. Finally, we monitor the structural similarity of the detected regions. Whenever a vehicle stalls or a traffic accident occurs, this would lead to a change in the structural similarity. Our contributions in this paper are as follows:
\begin{itemize}
    \item We propose an efficient framework for traffic video anomaly detection which is capable of running on the edge devices. 
    \item We propose a sequential anomaly detection algorithm which can significantly improve the robustness of the approach.
    \item We extensively test our algorithm on the 2021 AI City Challenge dataset without access to external data and yet perform comparatively well.
\end{itemize}

In summary, we propose an unsupervised framework which is capable of detecting anomalies, specifically stalled vehicles and crashes, in real-time. Our model has been tested on an NVIDIA Jetson AGX, which is a low-power System on Module (SOM) and can be deployed at the network edge. 
The experimental results reveal that with a resource-efficient framework, our approach is capable of performing competitively and is ranked fourth on the Track 4 test set of the AI City Challenge 2021, with an F1-score of 0.9157 at an RMSE of 8.4027. 

\section{Related Work}

While video anomaly detection has been extensively studied in the recent years, it still remains a challenging task. The existing approaches are mostly semi-supervised in nature, with models learning a notion of normality from the training videos and detecting activities that deviate from it. Early approaches focused on using handcrafted motion features such as histogram of oriented gradients (HOGs) \cite{chaudhry2009histograms,colque2016histograms,li2013anomaly}, Hidden Markov Models \cite{kratz2009anomaly,hospedales2009markov}, sparse coding \cite{zhao2011online,mo2013adaptive}, and appearance features \cite{cong2011sparse,li2013anomaly}. On the other hand, recent approaches have been completely dominated by deep learning based algorithms. These methods are based on specific scenes, e.g., they learn the nominal appearance, motion, and location features for each scene and then monitor the reconstruction error \cite{gong2019memorizing,hasan2016learning,luo2017revisit,nguyen2019anomaly,park2020learning} or the prediction error \cite{liu2018future,lee2019bman,dong2020dual,doshi2021online}. More recently, hybrid transfer learning based approaches have also been proposed \cite{doshi2020continual} which extract features using deep learning methods, use statistical approaches to detect anomalies, and is also capable of continual learning. 

On the other hand, there are also several supervised detection methods, which train on both nominal and anomalous videos. Particularly, \cite{sultani2018real} proposes using a deep multiple instance learning (MIL) approach to train on video-level annotated videos, in a weakly supervised manner. However, the main drawback of such methods is the difficulty in finding frame-level labeled, representative, and inclusive anomaly instances. Moreover, while they show a superior performance on similar anomalous events, they perform poorly when faced with unknown and novel anomaly types. 

In contrast to such approaches, traffic anomaly detection is more fine-grained and needs to be easy to generalize to different scenarios. To this end, \cite{fu2005similarity} studied detection of abnormal vehicle trajectories such as illegal U-turn. Particularly in the previous AI CITY Challenges \cite{xu2018dual,khorramshahi2019attention,biradar2019challenges,bai2019traffic,peri2020towards,doshi2020fast,li2020multi} used the background modeling method, which has achieved competitive results, to effectively eliminate the interference of the mobile vehicle, and obtains the location of the static region to analyze. Specifically, \cite{li2020multi} proposes a multi-granularity approach with pixel and box tracking to detect anomalies. They hand label the vehicles in the training data and then train a two-stage object detector on it, achieving the first rank. On the other hand, \cite{doshi2020fast} propose a fast unsupervised approach which uses k-nearest neighbors and K-means clustering for anomaly detection. Their approach uses a pretrained YOLO object detector, and achieves the second rank. However, all the above methods require detecting all the vehicles in the frames and then using a tracking algorithm to monitor their trajectory. This leads to considerable computational overhead and is not a feasible approach for deploying a vehicle anomaly detection system on the edge.  

\begin{figure*}[tbh]
\centering
\includegraphics[width=0.95\textwidth]{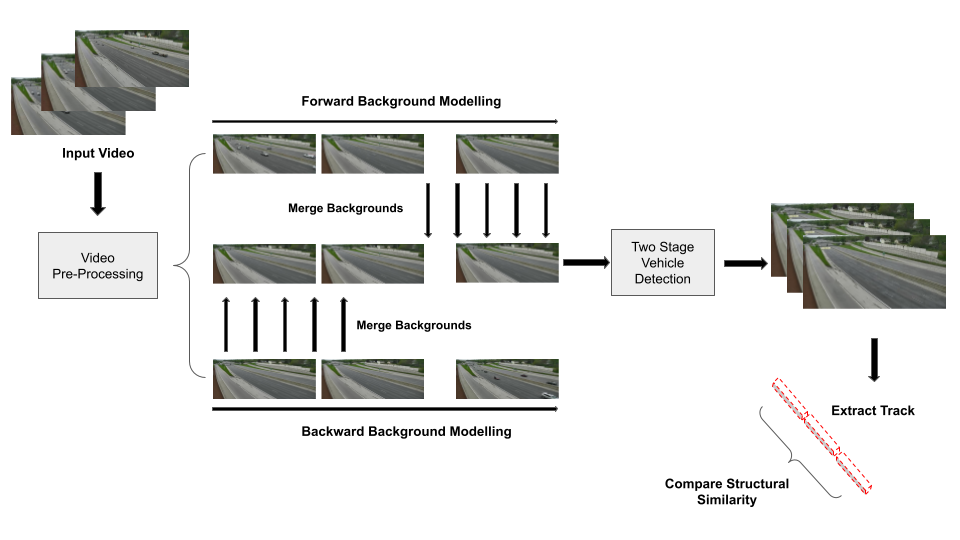}
\vspace{-2mm}
\caption{Proposed vehicle video anomaly detection system. We first extract the background in the forward and backward direction, and then only analyze the detected stationary vehicles.}
\label{f:system}

\end{figure*}
\section{Proposed Method}

While the primary task in this challenge is to detect vehicle anomalies as soon as possible, in general, it is also important to check the feasibility of the approach. While several recent algorithms have shown remarkable performance on several benchmark datasets, they also suffer from a significantly high computational overhead. Particularly, the state of the art algorithm proposed in \cite{bai2019traffic} trains the detection model on external datasets \cite{xu2018dual,zhu2018vision} and uses the computationally heavy ResNet-50 model which requires 311 ms per frame for vehicle detection on a reasonable GPU. In \cite{zhao2019unsupervised,wang2019anomaly}, a vehicle tracking approach is proposed. However, given the high number of vehicles detected in a traffic video, tracking each vehicle also becomes computationally inefficient.  While using a computationally expensive model certainly improves the performance, using it to detect vehicles every frame significantly increases the time required to process one video stream, thus defeating the purpose of a video anomaly detection system. Therefore, a trade-off between feasibility and performance is needed. 

To this end, we propose a more light-weight and intuitive approach. 
First, we employ a two-stage background modelling approach to remove all moving vehicles. We propose to focus only on the stationary objects that we see in the video, specifically cars and trucks. Hence, we train a two-stage object detector on the training videos using the labels from \cite{li2020multi}. Then, using a pretrained semantic segmentation model to remove the missclassified vehicles by removing those objects which occur outside the selected regions in the video. Finally, in the anomaly detection stage, given the region of interest, we locate the first instance where an anomalous vehicle is detected using a backtracking algorithm.

In the following subsections, we describe in detail the proposed approach for efficient anomaly detection. We begin by discussing the pre-processing and semantic segmentation stages. Then, we describe the background modelling and vehicle detection modules. Finally, we explain the anomaly detection algorithm, which enables timely and accurately detection of the onset of anomalies in the proposed framework. The entire algorithm is given in Algorithm \ref{algo1} and shown in Fig. \ref{f:system}. Additionally, we also propose a new framework by breaking down the video anomaly detection task into two separate problems, \emph{online} anomaly detection and \emph{offline} anomaly localization, along with a sequential anomaly detection approach, which we believe would help design more robust systems. 

\subsection{Pre-Processing and Semantic Segmentation}

The AI City Track 4 test dataset also includes videos which have several corrupted  or jittery frames. To deal with such videos, we employ a simple filter that measures the mean of the frame extracted every 30s and if it is close to zero, removes it. To extract background masks, we use a hierarchical multi-scale semantic segmentation model pretrained on the Cityscapes dataset \footnote{\url{https://github.com/NVIDIA/semantic-segmentation}}. The model uses HRNet-OCR as backbone and is more memory efficient than other approaches. It uses an attention based approach to combine multi-scale predictions. For the purpose of extracting masks, we consider roads and vehicles as a single label and all other classes as a different label.

\subsection{Background Modelling}

\begin{figure*}[tbh]
\centering
\includegraphics[width=0.95\textwidth]{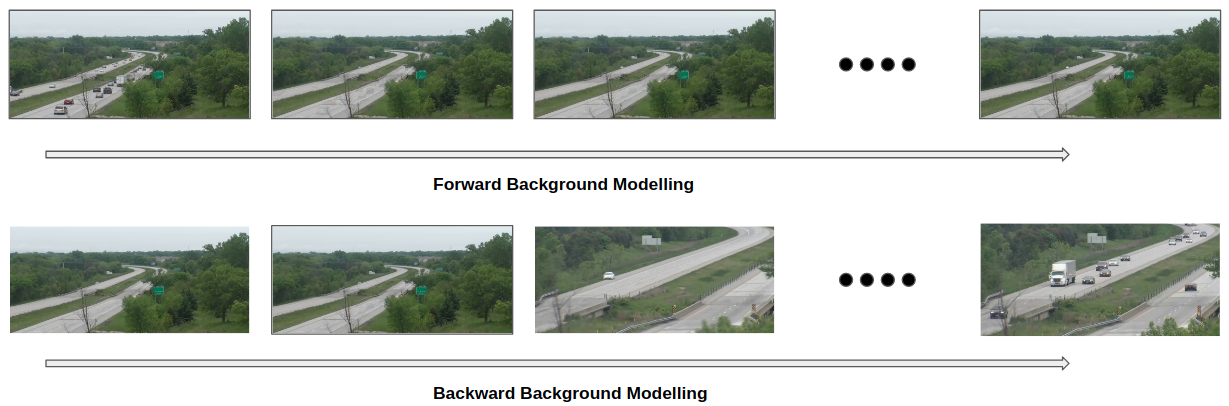}
\vspace{-2mm}
\caption{Comparison of forward background modelling and backward background modelling. We see that forward modelling suffers from visible vehicles in the initial images and backward modelling suffers from visible vehicles in the later images.}
\label{f:bb}
\end{figure*}

We hypothesize that stalled vehicles due to accidents or some mishap remain stationary for an extended period of time. Such vehicles, under decent lighting conditions would become a part of the background and make it easier for anomaly detection. To this end, for background modelling, we employ the MOG2 method \cite{zivkovic2004improved} in this paper. 
Since MOG2 is robust against image jittering and lighting variations, the extracted background masks do not have a lot of difference between them. Specifically, we set our update interval at 120 frames at 1 fps. While this reduces the computational cost, it takes longer for moving objects to completely disappear. However, for the purpose of the proposed algorithm, it is absolutely essential for all moving objects to be completely removed, or they could be sources of false positives. Hence, we extract the background in the forward and backward directions (Fig. \ref{f:bb}) and then merge them.

\subsection{Detection Model}

\begin{figure*}[tbh]
\minipage{0.24\textwidth}
 \includegraphics[width=\linewidth]{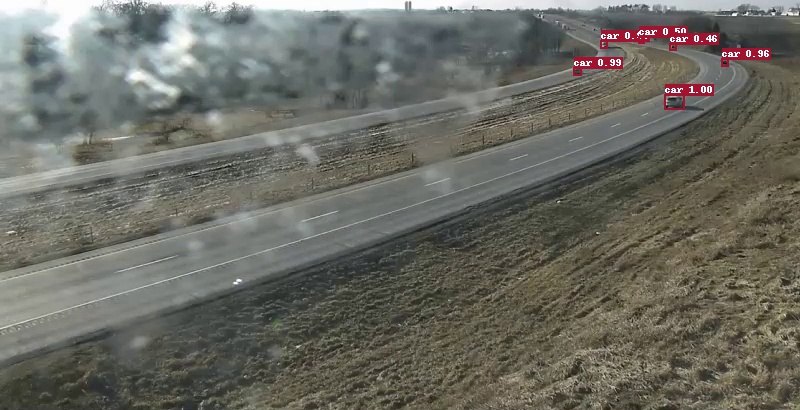}

\endminipage\hfill
\minipage{0.24\textwidth}
 \includegraphics[width=\linewidth]{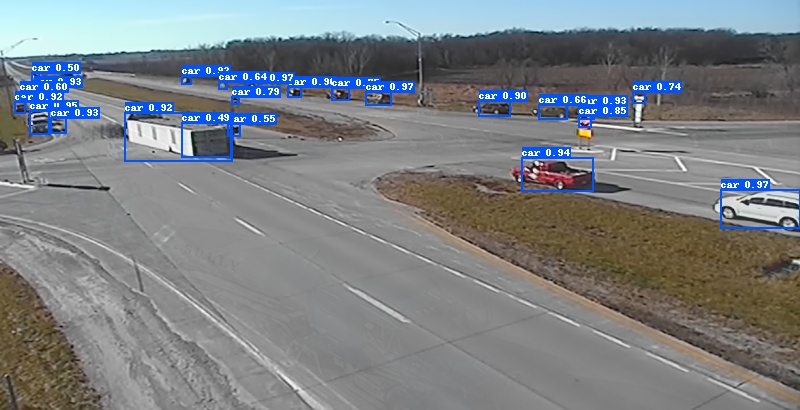}

\endminipage\hfill
\minipage{0.24\textwidth}%
 \includegraphics[width=\linewidth]{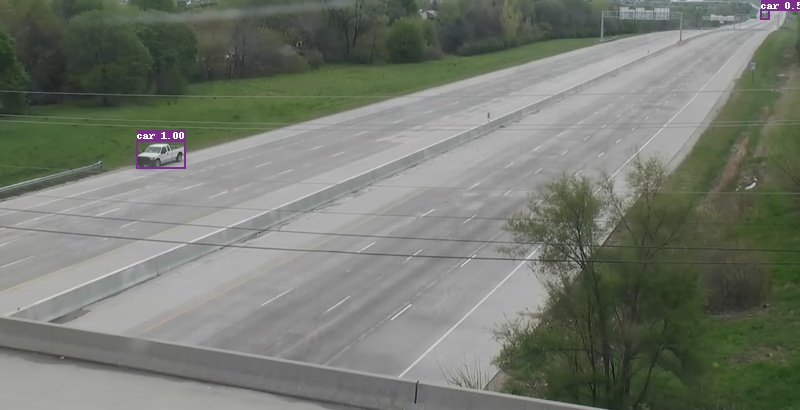}

\endminipage\hfill
\minipage{0.24\textwidth}%
 \includegraphics[width=\linewidth]{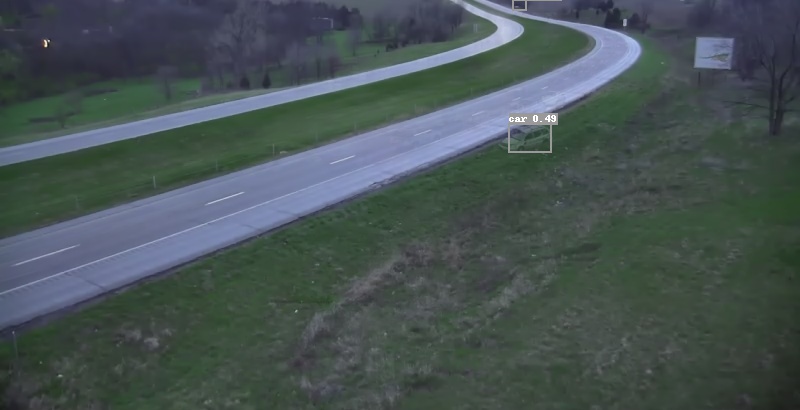}
\endminipage
\caption{Sample frames of detections.}
\label{f:det}
\end{figure*}

Object detection has received a lot of attention in recent years. Broadly, object detectors can be classified into single-stage and two-stage detectors. In single-stage object detectors such as YOLO (You Only Look Once) \cite{redmon2016you} and SSD (Single Shot Multibox Detector), the object detection task is treated as a simple regression problem, and directly output the bounding box coordinates. On the other hand, two-stage detectors such as Faster R-CNN \cite{ren2015faster} use a region proposal network first to generate regions of interest and then do object classification and bounding box regression. These methods are typically slower and take considerably longer, but are much better at detecting small objects. While single stage detectors are more efficient, we noticed that removing the false detections due to the lower accuracy accrues additional computational overhead, thus negating the advantage of using such detectors. To this end, following \cite{li2020multi}, we train a Faster R-CNN model which uses a Squeeze and Excitation Network (SENet) \cite{hu2018squeeze}, since they  generalize extremely well across different scenarios. SENet has a depth of 152 and uses a K-means clustering algorithm to cluster anchors, with the distance metric defined as:
\begin{equation}
    D(box,centroid) = 1 - IoU(box,centroid), \nonumber
\end{equation}
where $IoU$ denotes the intersection of union. 

We also leverage data augmentation to avoid overfilling and generalize the model to various lighting and weather conditions. Specifically, we use data augmentation techniques such as higher resolution, data flipping, and data cropping to improve the generalization capability of the algorithm. Since many of the vehicles are extremely small and difficult to detect, we also employ random cropping to learn on multiple scales. Each randomly cropped image is also resized to the recommended size of 1333x800. Moreover, to improve the detection recall, we also employ a feedback approach in combination to data flipping, which helps reduce the number of false positives. The images are randomly mirror flipped with a random probability of 0.5. The initial model is trained on the COCO dataset and then fine tuned using the AI City 2021 Track 4 training videos. We use the PaddlePaddle framework to train our model. The labelled training videos consist of frames extracted every four seconds from the training dataset and assigned bounding box level labels to the vehicles in images. We show some of the detected vehicles in Fig. \ref{f:det} 

Once object detection is performed for the entire video, we do some post-processing to eliminate stationary background objects detected as vehicle and slowly moving vehicles. Specifically, we map the center $(c_{xi}^t,c_{yi}^t)$ of the bounding box for an object $i$ detected at each time instance $t$ to a two dimensional plane. Then, for each point $(c_{xi}^t,c_{yi}^t)$, we compute the $k$-Nearest-Neighbor ($k$NN) distance $d_{xi,yi}^t(k)$ to its $k$ neighboring points. Specifically, we consider an object $i$ as misclassified if  
\begin{equation}
\label{eq:miss}
    d_{xi,yi}^t(k_1) \leq l_1,
\end{equation}
where $k_1 \gg l_1$, and as a slow moving vehicle if
\begin{equation}
\label{eq:slow}    
    d_{xi,yi}^t(k_2) \geq l_2,
\end{equation}
such that $k_2 \ll l_2$. 

\subsection{Backtracking Anomaly Detection}

\begin{figure*}[h]
\centering
\includegraphics[width=1\textwidth]{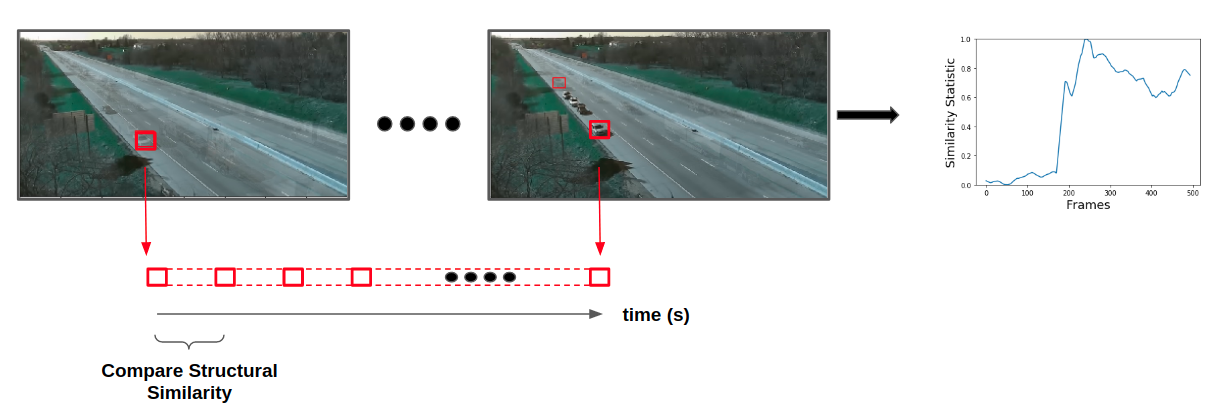}
\vspace{-3mm}
\caption{Backtracking anomaly detection pipeline for the proposed framework. We monitor the structural similarity (SSim) for each region of interest and decide to raise an alarm when the SSim crosses a threshold.}
\label{f:pipeline3}
\end{figure*}

For detecting the anomalies, we leverage the structural similarity (SSim) measure \cite{wang2004image}, which is a standard image comparison metric in Python libraries, for the frames within the detected region of interest, between the start of the video, $t=0$, and the instance when the stalled vehicle was detected, $t_k$. The motivation behind this is that in the absence of a stalled vehicle, the structural similarity with respect to an image with a stalled vehicle would remain close to zero, and would increase significantly when a vehicle appears. To remove increases caused due to noise, we apply a Savitzky-Golay filter to the similarity statistic. Specifically, we focus on whether the increase is \emph{persistant} over several frames or occurs only over a couple of frames. Finally, we declare $t$ as the onset time of the anomaly when the similarity statistic crosses the threshold. Fig. \ref{f:pipeline3} shows a case where a stopped car is successfully detected by our algorithm with minimum detection delay. In Algorithm \ref{algo1}, we summarize our entire pipeline.

\begin{algorithm}
\caption{Proposed anomaly detection algorithm}
\label{algo1}
\KwStage{\textbf{Preprocessing}}
\KwInput{$f^1$, $f^{120}$, \dots, $f^N$}
\KwOutput{ $(c_{xi}^1,c_{yi}^1)$, $(c_{xi}^{120},c_{yi}^{120})$, \dots, $(c_{xi}^N,c_{yi}^N)$}
\vspace{2mm}
\begin{algorithmic}[1]
   \FOR{$t = 1, 120, \ldots,N$}
        \STATE Obtain the averaged image $F_{avg}^t$ using MOG2.
        \STATE Determine bounding box for each detected object $i$.
        \STATE Remove overlapping boxes using NMS.
        \STATE Build set $C_{XY}$.
        \STATE Compute segmentation map $S$ using semantic segmentation.
    \ENDFOR 
\end{algorithmic}
\vspace{2mm}
\KwStage{\textbf{Candidate Selection}}
\KwInput{Set $C_{XY}$, segmentation map $S$}
\KwOutput{Centroid $(m_1,n_1), \dots, (m_K,n_K)$}
\begin{algorithmic}[1]
    \FOR{$t = 1, 100, \ldots,N$}
        \STATE Remove misclassified objects using Eq. \eqref{eq:miss}.
        \STATE Remove slow moving vehicles using Eq. \eqref{eq:slow}.
    \ENDFOR
    \STATE Select $K$ using elbow method.
    \IF {Centroid $(m_k,n_k)$ not in $S$}
        \STATE Remove $(m_k,n_k)$
    \ENDIF
\end{algorithmic}    
\KwStage{\textbf{Backtracking Anomaly detection}}
\KwInput{Anomaly detection time $t_{k}$ for centroid $k$, region of interest $(w_i^t,h_i^t)$}
\KwOutput{Anomaly onset time $\delta_t$}
\begin{algorithmic}[1]
    \FOR{$t = 1, 10, \ldots$}
        \FOR{$k = 1, \dots, K$}
            \IF {SSim ($ROI^t$,$ROI^{t_{k}}$) $>$ threshold)}
                \STATE Declare $t$ as anomaly onset time.
            \ENDIF            
        \ENDFOR
    \ENDFOR
\end{algorithmic}
\end{algorithm}

\section{Experiments}
\subsection{Experimental Setup}

The Track 4 dataset of the AI City Challenge 2021 consists of 100 training videos and 150 testing videos, each with an approximate length of 15 minutes. The videos are captured at a frame rate of 30 fps and a resolution of 800x410. The purpose of the challenge is to devise an algorithm that is capable of identifying all anomalies with minimum false alarms and detection delay. The anomalies are generally caused due to stalled vehicles or crashes. In contrast to previous years, the testing data is significantly more difficult since it consists of several videos with corrupted frames and noisy data. Hence, it is also very important to preprocess such videos since they can lead to false alarms. 

The evaluation for Track 4 had two major criteria, namely detection delay measured by the root mean square error (RMSE) and the detection performance measured by the $F_1$ score. Specifically, the final statistic was termed as $S_4$ and was computed as 
\begin{equation}
\label{eq:final}
    S_4 = F_1 (1 - NRMSE)
\end{equation}
where NRMSE is the normalized version of the root mean square error and is given by:
\begin{equation}
NRMSE = \frac{\min{\sqrt{\frac{1}{TP}\sum_{i=1}^{TP}((t_i^p-t_i^{gt})^2,300)}}}{300}    
\end{equation}
and $F_1$-score is given by:
\begin{equation}
    F_1 = \frac{2TP}{2TP+FN+FP}.
\end{equation}
The range of $S_4$ score is from 0 to 1, with 1 signifying the best performance that could be achieved. A detection was considered as a true positive if it was detected within 10 seconds of the true anomaly. 

\subsection{Implementation details}

In our implementations, the detection backbone network used is SENet-152, which is pretrained on the the COCO dataset and fine tuned on the labelled training data from the challenge. We use the suggested parameters, with stochastic gradient descent and a learning rate of 0.00125 and minibatch of 8. We train the model for 15K iterations, and the input images are resized to dimensions of 800x1333. The model utilizes 5 layer feature maps and use [16,32,64,128,256] as anchor boxes. During testing, the frames are extracted at 1 \emph{fps}. The Jetson AGX is initialized with NVIDIA JetPack 4.3 and compiled with PaddlePaddle. 

\subsection{Performance Evaluation}

\begin{table}[]
\centering
\begin{tabular}{|l|l|l|l|}
\hline
           & $F_1$                          & RMSE                        & $S_4$                          \\ \hline
Our Method & \multicolumn{1}{c|}{0.9157} & \multicolumn{1}{c|}{8.4027} & \multicolumn{1}{c|}{0.8900} \\ \hline
\end{tabular}%
\vspace{2mm}
\caption{Our performance}
\label{tab:my-table}
\end{table}

As shown in Table 1, we achieve an $F_1$ score of 0.9157 and an RMSE score of 8.4027. The final $S_4$ score computed using Eq. \ref{eq:final} is 0.8900, which placed us fourth in the challenge.  The fact that no external data was used shows the generalization capability of our proposed algorithm. In Table \ref{tab:leaderboard}, we show the results among all teams. 
The processing time of each step in our algorithm on the edge device NVIDIA Jetson Xavier is given in Table \ref{tab:timing}. The processing time is for each test video, which is approximately 15 minutes long. The object detection module is the most computationally expensive component; it requires about 14 minutes for each video. This can be drastically reduced by using a more efficient object detection model. The forward and backward background modelling take 2.5 minutes each and can be executed in parallel. The final anomaly detection requires 1.2 minutes, making the end-to-end algorithm real-time.   

\begin{table}[ht!]
\begin{center}

\begin{tabular}{cccc}
\hline
\hline
\multicolumn{4}{c}{\textbf{Leaderboard}} \\ \hline \hline
\textbf{Rank}   & \textbf{Id} & \textbf{Name} & \textbf{S4} \\ \hline \hline
1 & 76 & KillerSeven              & 0.9355 \\ \hline
2 & 158 & BD              & 0.9220 \\ \hline
3 & 92 & WHU\_IIP             & 0.9197 \\ \hline
\textbf{4} & \textbf{90}  & \textbf{SIS Lab}    & \textbf{0.8900} 
\\ \hline
5 & 153  & Titan Mizzou           & 0.5686 \\ \hline
6 & 48  & BUPT-MCPRL2       & 0.2890 \\ \hline
7 & 26  & Attention Please!              & 0.2184 \\ \hline
8 & 154 & Alchera & 0.1418 \\ \hline
9 & 12 & CET     & 0.1401 \\ \hline
\end{tabular}%
\end{center}
\vspace{2mm}
\caption{Result comparison on the Track 4 test set from the top 9 on the leaderboard. }
\label{tab:leaderboard}
\end{table}


\begin{table}[ht!]
\begin{center}

\begin{tabular}{cc}
\hline
\textbf{Component}   & \textbf{Processing Time (minutes)} \\ \hline \hline
Object Detection & 13.9\\ \hline
Forward BG Modelling & 2.5\\ \hline
Backward BG Modelling & 2.5\\ \hline
Backtracking & 1.2\\ \hline

\end{tabular}%
\end{center}
\vspace{2mm}
\caption{Processing time on NVIDIA Jetson Xavier AGX for a test video of ~15 minutes long.}
\label{tab:timing}
\end{table}

\section{Discussion}

Currently, the traffic video anomaly detection literature lacks a clear distinction between online anomalous event detection and offline anomalous frame localization. Particularly, online anomaly detection is essential for detecting emergencies in a timely manner, whereas offline anomaly localization is relevant in future video analysis. 
The $S_4$ evaluation metric used in the AI City Challenge considers the detection delay, but does not require the algorithm to be online. Specifically, it combines the event-level detection delay with the frame-level $F_1$ score. 
We believe that it is necessary to formally break down traffic video anomaly detection into two separate problems:
\begin{itemize}
\item {\it Online Detection:} Because of the time-sensitive nature of video anomaly detection, the proposed framework should be capable of real-time detection of anomalous events while minimizing the number of false alarms. 
\item {\it Offline Localization:} Once an anomalous event is reliably detected, an additional feature that the system can provide is temporally and spatially localizing the anomaly in frames and pixels, respectively. Since accuracy is the defining criterion here rather than quickness, this operation can be performed in an offline mode. 
\end{itemize}


\textbf{\textit{Problem Formulation:}} Given a video stream $F = \{f_1,f_2,\dots\}$, the system observes a new frame $\{f_t,x_t\}$ at each time instance $t$, comprising activity(s) $x_t$ drawn non i.i.d. from a complex and unknown distribution $\mathcal{X}$, which itself can undergo gradual or abrupt changes. For online anomalous event detection, the objective is to minimize the average detection delay while maintaining an acceptable false alarm/positive rate. Hence, we formulate this in terms of quickest change detection framework \cite{basseville1993detection}, which is tailored for this objective: 
\begin{align*}
\mathcal{X} = \mathcal{X}_0 ~\text{for}~ t<\tau, ~\text{and}~ \mathcal{X} \not= \mathcal{X}_0 ~\text{for}~ t \ge \tau.
\end{align*}
where $\tau$ is the instance when the anomaly begins. In this formulation, the objective is to quickly and reliably detect the time instance $\tau$. This is fundamentally different from frame localization, in which the objective is to classify the frames as nominal or anomalous without regard to continuity of video stream and anomalous event.

After detection of an anomalous event, given a video segment, the offline anomaly localization problem can be formulated as a binary hypothesis test for each frame. The vast majority of existing works implicitly pursue this offline localization problem for video anomaly detection. 

\textit{\textbf{Performance Metric for Online Detection:}} 
Since the online detection of anomalous events in streaming video is not rigorously studied in the literature, there is a need for a suitable performance metric to evaluate algorithms addressing this problem. While the commonly used frame-level AUC (area under the ROC curve, borrowed from binary hypothesis testing) might be a suitable metric for localizing the anomaly in video frames, it ignores the temporal nature of videos and fails to capture the dynamics of detection results, e.g., a detector that detects the later half of an event holds the same AUC as the detector that detects every other frame. 
To avoid cases where a detector might always raise an alarm, the false alarm rate also needs to be monitored. 
We here present a new performance metric called Average Precision Delay (APD), which is based on average detection delay and alarm precision: 
\begin{equation}
\label{eq:met}
    \text{APD} = \int_{0}^{1} P(\alpha) ~\text{d}\alpha,
\end{equation}
where $\alpha$ denotes the normalized average detection delay, and $P$ denotes the precision. The average detection delay is normalized by the largest possible delay either defined by a performance requirement or the length of natural cuts in the video stream. Here, precision is defined as the ratio of number of true alarms to the number of all alarms. 
APD $\in [0,1]$ measures the area under the precision-delay curve to give a comprehensive performance metric for online video anomaly detection. A highly successful algorithm with an APD value close to $1$ must have \emph{high reliability} in its alarms (i.e., precision close to $1$) and \emph{low delay}. 

\textit{\textbf{Sequential Anomaly Detection:}} Due to the temporal nature of traffic videos, we propose a nonparametric sequential anomaly detection algorithm which aims to achieve high alarm precision and low detection delay. The proposed algorithm leverages the computed structural similarity SSim $e_t$ for each extracted frame $f_t$ in the training dataset. Once we compute the structural similarity, we normalize them between $[0,1]$ and then extract the nonzero values and build a set $S_{train}$. Then, for a significance level, e.g., $\alpha=0.05$, the ($1-\alpha$)th percentile $\gamma$ of $S_{train}$ is computed. This is used as a baseline statistic to remove the nominal structural similarity. For online testing for anomalous events, following the statistic update rule of the minimax-optimum CUSUM algorithm \cite{basseville1993detection}, we accumulate the anomaly evidence $e_t-\gamma$ to obtain the test statistic $s_t$:
\begin{equation}
    \label{eq:stat}
    s_t = \max\{0,s_{t-1} + e_t - \gamma\}.
\end{equation}
 Finally, the test continues until the accumulated anomaly evidence $s_t$ exceeds a predetermined threshold $h$, which controls the trade-off between the conflicting objectives, high precision and low delay. Once an anomalous event is detected at time $T=\min\{t: s_t \ge h\}$, we find the point where $s_t$ starts continuously decreasing, say $T+M$, and compare the anomaly evidence $e_t, t \in [T,T+M]$ for each suspicious frame with a threshold $g$ for anomaly localization. The threshold $g$ also determines the trade-off between the true positive and false positive rates in localization. 

\begin{figure}
\includegraphics[width=\linewidth]{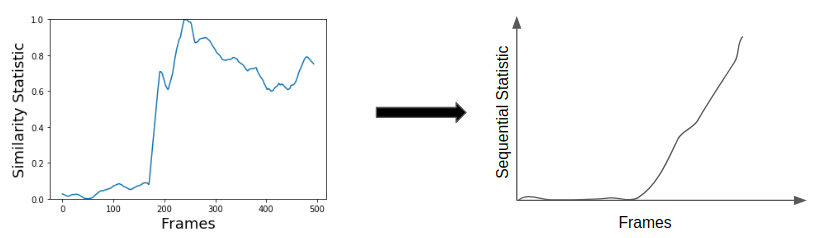}
\caption{The advantage of sequential anomaly detection over a single-shot detector. It is seen that a sequential detector can significantly reduce the number of false alarms.}
\label{f:sequen}
\end{figure}

To illustrate the significance of the proposed sequential detection method, we compare the structural similarity computed to the sequential algorithm. As shown in Fig. \ref{f:sequen}, the proposed sequential statistic handles noisy evidence by integrating recent evidence over time. On the other hand, the instantaneous anomaly evidence is more prone to false alarms since it only considers the noisy evidence available at the current time to decide.
\section{Conclusion}

In this work, we proposed an efficient solution for traffic video anomaly detection. The proposed solution is able to perform competitively in the 2021 AI City Challenge and run at an edge device such as NVIDIA Jetson Xavier. 
We also highlighted key shortcomings in the existing problem formulation and proposed a new framework that addresses them. Since the ground truth was unavailable, we were unable to evaluate the performance of our model on the proposed online event detection metric, but we hope that it can be helpful for future algorithm design. Our future work would include density estimation for the $K$-means algorithm and a continual learning based model capable of learning different type of anomalies. 

{\small
\bibliographystyle{ieee_fullname}
\bibliography{Ref.bib}
}

\end{document}